\pdfoutput=1
\documentclass[11pt]{article}

\usepackage{EMNLP2022}

\usepackage{times}
\usepackage{latexsym}
\usepackage{graphicx,amsmath,amsthm,amsfonts,epsfig,bbding}
\usepackage{color}
\usepackage{amssymb}
\usepackage{mathptmx}
\usepackage{subfigure}
\usepackage{hyperref}
\usepackage{threeparttable}
\DeclareMathAlphabet{\mathcal}{OMS}{cmsy}{m}{n}
\usepackage{booktabs}

\usepackage[T1]{fontenc}
\usepackage[utf8]{inputenc}

\usepackage{microtype}
\usepackage{graphicx} 
\usepackage{inconsolata}

\title{Contrastive Learning Subspace for Text Clustering}

\author{Qian Yong \\ Alibaba Group \\  Hangzhou, China \\ sarahyongq@gmail.com
        \AND
        Chen Chen \\ Soochow University\\ Suzhou, China \\ cc336655@gmail.com \And
        Xiabing Zhou \\ Soochow University \\ Suzhou, China \\ zhouxiabing@suda.edu.cn }
\begin{document}
\maketitle
\begin{abstract}

% Contrastive learning is an unsupervised learning method to get effective representations for clustering tasks. Existing contrastive learning methods mainly focus on the instance-wise relationships, which can greatly improve semantic similarity tasks. However, the cluster-wise relationships are much more important for clustering tasks. Therefore, we propose a method called Contrastive Learning Based on Subspace (CLBS), which constructs virtual positive samples based on the self-expression theory of subspace clustering, so that the model forms different subspaces during contrastive learning and captures the cluster-wise relationships. Experiments show that our contrastive learning method has achieved better results on clustering tasks with lower cost of positive sample construction.

%%% Modified version:

Contrastive learning has been frequently investigated to learn effective representations for text clustering tasks. While existing contrastive learning-based text clustering methods only focus on modeling instance-wise semantic similarity relationships, they ignore contextual information and underlying relationships among all instances that needs to be clustered. In this paper, we propose a novel text clustering approach called Subspace Contrastive Learning (SCL) which models cluster-wise relationships among instances. Specifically, the proposed SCL consists of two main modules: (1) a self-expressive module that constructs virtual positive samples and (2) a contrastive learning module that further learns a discriminative subspace to capture task-specific cluster-wise relationships among texts. Experimental results show that the proposed SCL method not only has achieved superior results on multiple task clustering datasets but also has less complexity in positive sample construction.

\end{abstract}

\section{Introduction}

%%% Clustering is always a good choice when categorizing unlabeled data or difficult-to-label data. The conventional clustering method is usually a two-step mode. Taking text clustering as an example, texts are vectorized through language models such as word2vec firstly, and then the text vectors are clustered based on clustering algorithms such as KMeans. However, this two-step mode means that the algorithm can not optimize the distance between samples while clustering. Therefore, in addition to the clustering algorithm itself, the representations of texts also determine the upper limit of the clustering accuracy.

%%% Modified version:

Clustering is an typical unsupervised learning technique that has been widely employed to categorize unlabeled data or difficult-to-label data. Traditional text clustering methods for natural language processing (NLP) usually consists of two stages: text representation learning and categorization, i.e., texts are firstly represented as a set of feature vectors, which are then categorized into several clusters in a unsupervised manner. For example, previous approaches frequently employ word2vector \cite{mikolov2013efficient} or sentence-BERT \cite{DBLP:journals/corr/abs-1908-10084} to encode original texts to representations and then apply K-means to obtain the final text clusters.

% Text representations based on pre-training language models usually contain more semantic information and have impressive performance on tasks such as text classification. Researchers \cite{DBLP:journals/corr/abs-1908-10084, DBLP:journals/corr/abs-2103-15316, li2020emnlp, gao2021simcse} proposed different methods to improve the performance of BERT-based embeddings on similarity tasks. But they are weak on a series of similarity tasks such as semantic similarity and text clustering. 
% % 对比学习是一种较好地提高语义相似度任务的方法
% % 对比学习在其中的简要运用，但无法准确获取cluster wise 信息
% % SCCL与CC方法可以获取cluster wise信息，但需要与聚类方法配合，预先设定好聚类数目

%%% Modified version:

% However, such two-stage approaches means that the algorithm can not optimize the distance between samples while clustering.
% %%% 传统的聚类在做表征的时候没考虑到聚类

Consequently, in addition to the clustering algorithm itself, the representations of texts is also a key factor determining the clustering performance. In other words, text representation learning is crucial for text analysis tasks, where well designed representation learning models can usually extract representations containing more task-specific semantic information which can largely enhance the performance of various text tasks such as text classification \cite{LillebergZZ15}, text clustering \cite{XuWTXZWH15}, etc. Although previous studies \cite{DBLP:journals/corr/abs-1908-10084, DBLP:journals/corr/abs-2103-15316, li2020emnlp,gao2021simcse} proposed various schemes to enhance the BERT-based representations, they usually fail to clearly improve the performance of text clustering tasks.

% %%% 增添一下目前text representation learning 的方法介绍和缺陷
% Researchers \cite{DBLP:journals/corr/abs-1908-10084, DBLP:journals/corr/abs-2103-15316, li2020emnlp, gao2021simcse} proposed different methods to improve the performance of BERT-based embeddings on similarity tasks. 

% But they are weak on a series of similarity tasks such as semantic similarity and text clustering. 
% %%% 表征学习，现有two-stage: instance-wise, 我们用的是group-wise

\begin{figure}[t]
	\centering
\includegraphics[width=230px]{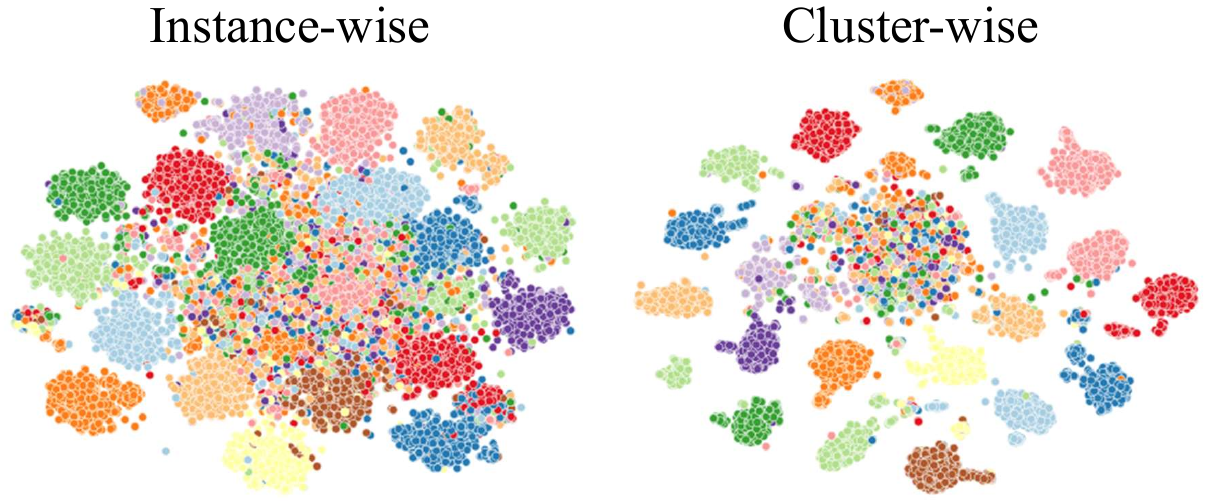}
	\caption{\label{figure-scatter} TSNE visualization of the embedding space
learned on StackOverflows using Sentence Transformer as backbone. Each
color indicates a ground truth semantic category. The boundaries between clusters based on cluster-wise contrastive learning are more clear than instance-wise methods.}
\end{figure}

% The main idea is to optimize language models based on contrastive learning.
% Contrastive learning is an instance-wise self-supervised learning method, which aims to learn the effective representation by pulling together samples augmented from the same instance and pushing apart those less related ones. Therefore, similar samples are gathered together and different types of samples will distribute on the hypersphere as soon as possible. For example, simCSE \cite{gao2021simcse} achieved excellent results through contrastive learning with data enhancement in the form of dropout twice. However, for clustering tasks, we need to judge whether the formation of sample clusters is reasonable from the perspective of cluster-wise and whether the boundaries between clusters are clear enough. 

% Some improvement methods combine contrastive learning with clustering methods directly to get the cluster-wise information. SCCL \cite{zhang2021supporting} combines contrastive learning with DEC \cite{GuoGLY17}, a deep clustering algorithm, to assign clusters to samples while contrastive learning, which greatly improves the quality of cluster formation. Contrastive clustering \cite{li2021contrastive} takes the rows of the feature matrix as soft labels of instances and the columns as cluster representations. However, those methods both need the key prior knowledge: the number of categories, and restrict the scenarios. Besides, most contrastive methods bear the high cost of positive sample construction and need to encode these samples twice. 

%%% Modified version

To learn more discriminative and task-specific text representations, recent methods have frequently extended contrastive learning algorithms to text clustering tasks. Specifically, contrastive learning enforces models to learn a subspace where samples belong to the same category have very similar representations, while the representations of samples belong to different categories have long distances, which has achieved superior performances on text clustering tasks (e.g., simCSE \cite{gao2021simcse}, SCCL \cite{zhang2021supporting}, Li et al. \cite{li2021contrastive}). However, all of such methods require a key prior knowledge: the number of categories, which may not available in many real-world scenarios, and thus limits their usage. In addition, most contrastive learning methods suffer from a high cost of positive sample construction and need to encode these samples twice.

To tackle the aforementioned challenges, we propose a novel Subspace Contrastive Learning (SCL) approach for text clustering, i.e., the proposed SCL utilizes contrastive learning to learn an optimal deep subspace model. The hypothesis of this paper is that each sample can be represented by a linear combination of several features lying in a set of latent subspaces. Specifically, our method aims to capture the cluster-wise relationships with the lower cost of positive sample construction while not requiring the number of categories. We treat the self-expression as an augmented method to construct the virtual positive samples for contrastive learning, which not only portrays the cluster-wise relationship, but also avoids complex two-stage text augmentation for existing NLP models, i.e., while existing contrastive learning approaches \cite{} would individually feed each original sample and its corresponding positive sample to the encoder, our SCL construct postive sample pairs, which can jointly feed them to the encoder. Figure \ref{figure-scatter} compares distribution of features learned with instance-wise contrastive learning and cluster-wise contrastive learning. The main contributions of this paper are summarized as follows:
\begin{itemize}
    \item We propose a cluster-wise contrastive learning approach which allows models to learn the cluster-wise relationship between input samples through subspace modeling. To the best of our knowledge, this is the first cluster-wise contrastive learning approach for text clustering.
    
    \item We propose a new self-expressive method for constructing virtual positive samples, which is not only faster but also has better generalization capability when combined with shallow networks compared with other text augmentation techniques.
    
    \item Our comprehensive experiments on text clustering task results show that our method outperforms the State-of-the-art method on multiple datasets.
\end{itemize}

\section{Related Works}

\subsection{Contrastive Learning}

Contrastive Learning is an unsupervised learning method to learn effective representations, which has been widely applied in various visual representation learning \cite{oord2018representation,he2020momentum}, 
representation learning \cite{su2020dialogue,lin2020world} and text representation learning\cite{iter2020pretraining, ding2021prototypical}. The main idea of the contrastive learning is to enforce models to encode similar representations from samples augmented from the same instance and enlarge the distance among less related ones, so that similar samples are gathered together and different types of samples will distribute on the hypersphere as far as possible. 

As a result, contrastive learning has frequently improved the performances of various downstream works. For example, semantic similarity tasks employ contrastive learning over pre-training language models to get more similarity-sensitive sentence representations \cite{gao2021simcse, wu2021esimcse} and clustering tasks learn a clustering object and a contrastive object jointly \cite{zhang2021supporting, li2021contrastive} to relieve the overlap problem across categories before clustering starts. However, applying contrastive learning to NLP tasks sometimes faces the data augmentation problem which is much more difficult to tackle compared with CV field. Thus, text augmentation for contrastive learning has attracted a lot of interest. Zhang et al. \citet{zhang2021supporting} proposed the contextual augmenter which leverages the pre-trained transformers to find top-n suitable words of the input text for insertion or substitution. Gao et al. \citet{gao2021simcse} takes dropout as a minimal data augmentation method, and passes the same input sentence to a pre-trained Transformer encoder twice. Wu et al. \citet{wu2021esimcse} further addressed the length bias of simCSE by repeating words from the input sentence.

\begin{figure*}[ht]
	\centering
	\includegraphics[width=360px]{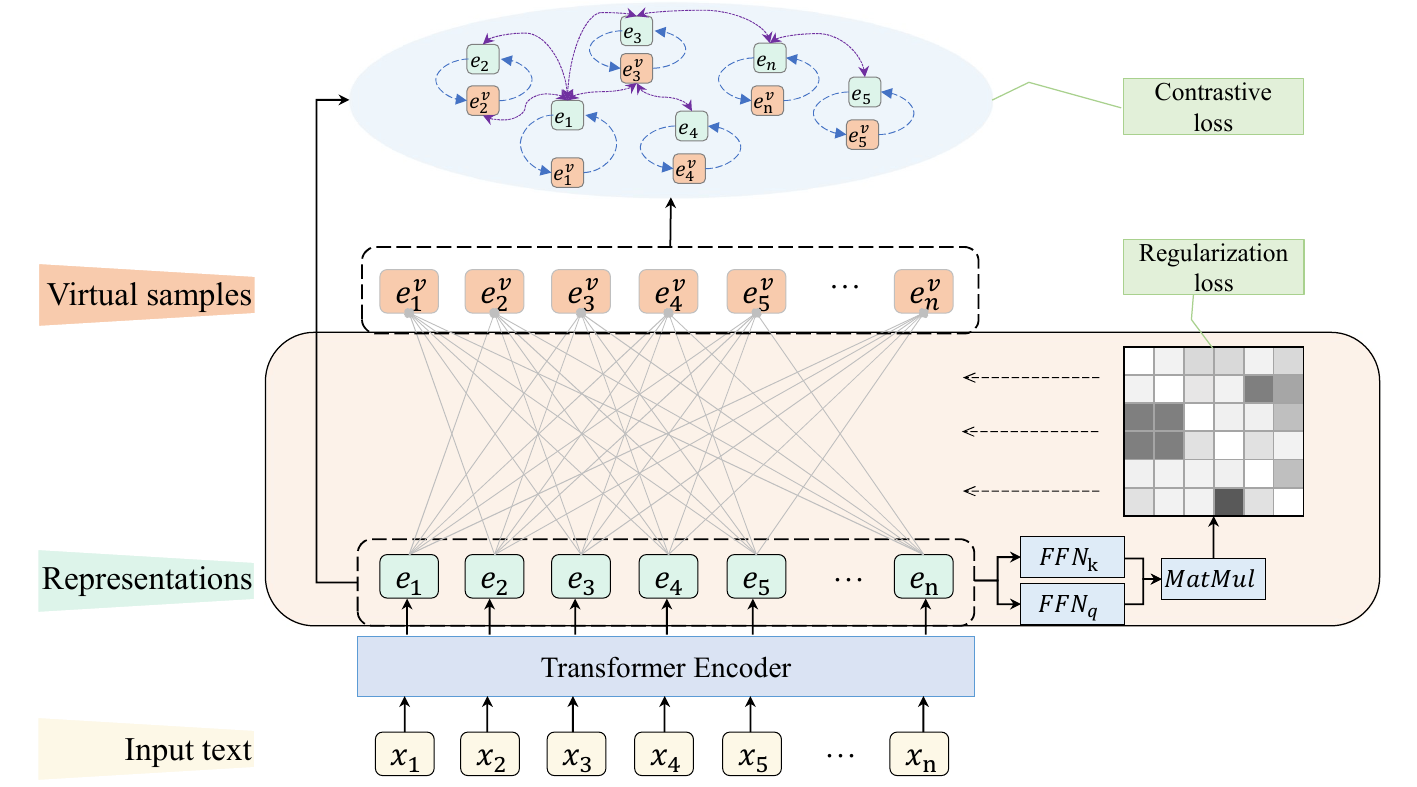}
	\caption{This is architecture of the method we proposed and take the green sample as an example. The inputs are encoded as features firstly and then generate virtual samples to form positive pairs (red sample) and negative pairs (purple and yellow samples). Contrastive learning will be applied to pull together positive pairs and push apart negative pairs.}
	\label{clbs_model}
\end{figure*}

\subsection{Subspace Clustering}

Subspace clustering is an unsupervised learning approach, which aims at finding out underlying subspaces to cluster data points from the same subspace. It has attracted lots of interests in computer vision tasks, such as image clustering \cite{ho2003clustering, elhamifar2013sparse}, motion segmentation \cite{chen2009spectral, costeira1998multibody} and multiview clustering \cite{peng2019comic}. In these approaches, a crucial step is to learn an affinity matrix $A$ where $A_{ij}$ represents the correlation coefficient between the $i$th and $j$th sample, which is then used for spectral clustering \cite{ng2002spectral} in order to obtain the final clustering results. 

Consequently, many subspace clustering studies have been devoted to applying  self-expression based methods \cite{liu2010robust, lu2012robust,wang2013provable,li2015structured} to produce high quality and robust affinity matrices for deep learning. Here, self-expression is defined as expressing each data point as a linear combination of other points in the same subspace and the combination coefficients refer to the affinity matrix. In addition, some regularization functions such as L1 \cite{2009Sparse} or L2 \cite{favaro2011closed} have been applied to ensure that
nonzero coefficients $A_{ij}$ occur only between $i$th and $j$th sample lying in the same subspace \cite{li2016structured}. To compute self-expressive more effectively, scalable subspace clustering methods are proposed to reduce the $N \times N$ computational complexity by either subsampling \cite{peng2013scalable,traganitis2017sketched} or decomposing a large-scale optimization problem into a sequence of small scale problems \cite{you2016scalable, dyer2013greedy}. Zhang et al. \citet{zhang2021learning} employs self-attention to learn the self-expressive representation and get competitive results.

\section{The proposed approach}

% We aim at learning cluster-aware sentence representations via cluster-wise contrastive learning, which constructs virtual positive samples from self-expressive mechanism. As illustrated in Figure \ref{figure-main}, our model consists of three components. The first one is an encoder initialized from language model, which needs to be adapted with downstream clustering tasks. The second one is the self-expressive layer which can be considered as a self-attention layer to construct virtual positive samples. The last one is the optimizer objects includes contrastive object and regularization object. We will highlight the details of Self-expressive Layer and Cluster-wise Contrastive Learning bellow. 

%%% modified version

In this section, we propose a novel cluster-wise contrastive learning algorithm called Subspace Contrastive Learning (SCL) for learning cluster-aware text representations, which generates virtual positive samples from the self-expressive mechanism. As illustrated in Figure \ref{clbs_model}, our approach consists of two main modules: (1) a pre-trained transformer-based encoder that first vectorizes $N$ input sentences $X = [x_1, x_2,\cdots,x_N] \in \mathbb{R}^{N \times D}$, and encodes them as a set of latent representations $E = [e_1, e_2,\cdots,e_N]$, which would be further adapted to downstream clustering tasks; and (2) a self-expressive module that constructs virtual positive samples (Sec. \ref{Cluster-wise}), where we provide the object function for optimizing our model in a constrastive manner.

%%% 等图画好后再继续修改
\subsection{Self-expressive Module}
\label{self-expressive}

%%% 就是个attention 层

% Firstly, the minibatch input sentences are vectorized as $X = [x_1, x_2,\cdots,x_N] \in R^{N \times D}$ with the encoder and meaning pooling. And then, we employ the self-expressive layer to construct positive samples. Self-expressive methods for subspace are based on solving for each $j \in {1,2,\cdots,N }$ and optimization problem of the form:
% \begin{equation}
% \min _{\left\{c_{i j}\right\}_{i \neq j}} \frac{\gamma}{2}\left\|\boldsymbol{x}_{j}-\sum_{i \neq j} A_{i j} \boldsymbol{x}_{i}\right\|_{2}^{2}+\sum_{i \neq j} r\left(A_{i j}\right)
% \end{equation}
% where $A_{ij}$ is the coefficient between $i$th and $j$th sample. The first part aims at maximizing the similarity between origin representations $x_j$ and virtual representations $\hat{x_j}$ constructed from the linear combination of other data points. The second part is a regularization object on the coefficient matrix $A$ so that the coefficient matrix can become sparse or low-rank, and the most idealistic situation is that nonzero coefficients $A_{ij}$ occur only between $i$th and $j$th sample lying in the same subspace. This part can be abbreviated as $\mathcal{L}_{regularization}$.

%%% Modified version
We propose a novel self-expressive module to efficiently generate virtual positive samples for text augmentation. Specifically, the self-expressive module conducts an attention operation. It first learns query matrix $E^q$ and key matrix $E^k$ from the latent representations $E = [e_1, e_2,\cdots,e_N]$, and then conducts matrix multiplication to obtain affinity matrix $A \in \mathbb{R}^{N \times N}$. Finally, virtual positive samples $E^v$ can be obtained by further conducting matrix multiplication between the affinity matrix $A$ and the latent representations $E$ produced from the encoder. This process can be formulated as:
\begin{equation}
\begin{split}
    & A = E^q E^k \\
    & E^v = (A-I)E
\end{split}
\end{equation}
where we exclude cases that $i=j$

In particular, $A$ is optimized in the form of: 
\begin{equation}
\begin{split}
    & \min _{\left\{c_{i j}\right\}_{i \neq j}} \frac{\gamma}{2}\left\|\boldsymbol{e}_{j}-e^v_j\right\|_{2}^{2}+\sum_{i \neq j} r\left(A_{i j}\right) \\
    & e^v_j=\sum_{i \neq j} A_{i j} \boldsymbol{e}_{i}
\end{split}
\end{equation}
where $j \in {1,2,\cdots,N }$; $A_{ij}$ is the coefficient between $i$th and $j$th sample. Specifically, the first part $\left\|\boldsymbol{e}_{j}-\sum_{i \neq j} A_{i j} \boldsymbol{e}_{i}\right\|_{2}^{2}$ aims at maximizing the similarity between the original latent representations $e_j$ and the generated virtual samples $e^v_j$. The second part $\sum_{i \neq j} r\left(A_{i j}\right)$ is a regularization object to allow the coefficient matrix becoming sparse or low-rank, where the most idealistic situation is that nonzero coefficients $A_{ij}$ only occur when $i$th and $j$th sample belong to the same subspace. This part can be abbreviated as $\mathcal{L}_{regularization}$.

% We then conduct a soft attention to $X$ to get the coefficients matrix $A$ just like self-attention dose, and apply an activation function to filter those noisy connections:
% \begin{equation}
%     A = \mathcal{T}_{b}(X W_q \times X^T W_k^T),
% \end{equation}
% \begin{equation}
% \mathcal{T}_{b}(t):=\operatorname{sgn}(t) \max (0,|t|-b).
% \end{equation}
% In the above, $W_q$ and $W_k$ refer to learnable parameters of query and key respectively. $\mathcal{T}_{b}(\cdot)$ is a learnable soft thresholding operator Where $b$ is the learnable threshold to filter those coefficients bellow threshold.

% Excluding $i=j$, our virtual positive samples can be constructed as following:
% \begin{equation}
%     \tilde{X} = (A-I) \times X.
% \end{equation}

\subsection{Cluster-wise Contrastive Learning}
\label{Cluster-wise}

To allow our model to generate cluster-aware virtual samples, we propose a novel cluster-wise constrastive learning loss. This loss aims to enforce the model to learn subspaces where the distance between each pair of real sample and virtual positive sample to be minimized while the distance of negative pairs are maximized. 

Given a batch of training samples, we treat each latent representation $x_i$ and corresponding virtual positive representations $e^i$ generated by our self-expressive module as a pair of positive pair (i.e., N pairs of positive pairs). Meanwhile, we pair each latent representation $x_i$ with the other $N-1$ samples (i.e., $e^1, e^2, \cdots, e^{i-1}, e^{i+1}, \cdots$) and $N-1$ generated virtual samples as negative pairs. This would totally result in $2N-2$ negative pairs. We then propose a cluster-wise constrastive learning loss function as follows:
\begin{equation}
\ell_{i}^{I}=-\log \frac{e^{\operatorname{sim}({x}_{i}, e_{i}) / \tau}}{\sum_{j=1}^{N} 
(e^{\operatorname{sim}({x}_{i}, {x}_{j}) / \tau} +
e^{\operatorname{sim}({x}_{i}, e_{j})/\tau})}.
\end{equation}
where $t$ is an initial temperature parameter and is negatively correlated with the averaged similarity between positive pairs in a batch of inputs. Consequently, the punishment on negative pairs will be relieved until the self-expressive representations become credible. This part can be abbreviated as $\mathcal{L}_{contrastive}$.
where $sim(\cdot)$ is the cosine similarity function and $\tau$ denotes the temperature parameter which controls the punishment intensity on negative samples, i.e., smaller $\tau$ leads more punishment. However, the virtual positive representations are generated and even have lower confidence compared with some real "negative" samples during the initial training epochs. Therefore, we set $\tau$ automatically adjusted with the virtual positive samples defined as:
\begin{equation}
    \tau = t/{\text{average}(sim(x_i, e_v))},
\end{equation}
where $t$ is an initial temperature parameter and is negatively correlated with the averaged similarity between positive pairs in a batch of inputs. Consequently, the punishment on negative pairs will be reduced until the self-expressive module can generate reliable virtual samples, i.e., the postive pairs and negative pairs can be easily distinguished. This part can be abbreviated as $\mathcal{L}_{contrastive}$.

In summary, we jointly optimize the encoder and self-expressive module in a contrastive learning manner as:
\begin{equation}
    \mathcal{L} = \lambda_{cl} \mathcal{L}_{contrastive} + \lambda_{reg} \mathcal{L}_{regularization}.
\end{equation}

% \subsection{Overall objective}
% \label{Objective function}

% We optimize the self-expressive object and contrastive object jointly. From all of the three loss objects, we are surprised to find that recovering the original representations from the self-expressive layer just equals to maximizing the similarities between positive pairs. Thus, our optimizer objects can be reduced to the following:
% \begin{equation}
%     \mathcal{L} = \lambda_{cl} \mathcal{L}_{contrastive} + \lambda_{reg} \mathcal{L}_{regularization}.
% \end{equation}

\section{Experiments}

In this section, we first introduce the employed datasets and implementation details in Sec. \ref{Datasets and implementation}. Then, we report a series of experimental results to demonstrate the effectiveness and various aspects of the proposed SCL in Sec. \ref{Results and analysis}.

and the baseline methods employed for comparison. We then report the experimental results conducted from different perspectives and analyze the effectiveness of the proposed model with different factors.

\subsection{Datasets and implementation details}
\label{Datasets and implementation}

\begin{table}[t]
\begin{center}
\begin{tabular}{c|cccc}
\hline Dataset & $|V|$ & \multicolumn{2}{c}{ Doc } &  Cls\\
& & $N^{D}$ & Len & $N^{C}$ \\
\hline AgNews & $21 \mathrm{~K}$ & 8000 & 23 & 4 \\
StackOverflow & $15 \mathrm{~K}$ & 20000 & 8 & 20 \\
SearchSnippets & $31 \mathrm{~K}$ & 12340 & 18 & 8 \\
GooglenewsTS & $20 \mathrm{~K}$ & 11109 & 28 & 152 \\
GooglenewsS & $18 \mathrm{~K}$ & 11109 & 22 & 152\\
GooglenewsT & $8 \mathrm{~K}$ & 11109 & 6 & 152 \\
Tweet & $5 \mathrm{~K}$ & 2472 & 8 & 89 \\
\hline

\end{tabular}
\caption{\centering Dataset statistics. $|V|$: the vocabulary size;
$N^D$: number of short text documents; Len: average number of words in each document; $N^C$ number of clusters}
\label{table-dataset} \centering
\end{center}
\end{table}

\begin{table*}[t]
\begin{center}
\begin{tabular}{lcccccccc} 
\hline
& \multicolumn{2}{c}{ SearchSnippets } & \multicolumn{2}{c}{ StackOverflow } & \multicolumn{2}{c}{ AgNews } & \multicolumn{2}{c}{ Tweet } \\
& ACC & NMI & ACC & NMI & ACC & NMI & ACC & NMI \\
% \cline { 2 - 9 } 
\hline
SBERT & $55.6$ & $31.8$ & $56.7$ & $46.9$ & $67.7$ & $32.6$ & $51.8$ & $77.7$ \\
WordNet & $74.3$ & $54.5$ & $71.6$ & $69.3$ & $83.1$ & $57.8$ & $60.0$ & $82.0$ \\
Back-Trans & $ 78.0 $ & $64.8$ & $77.3$ & $\bf75.5$ & $82.7$ & $57.5$ & $59.7$ & $82.3$ \\
Contextual & $74.7$ & $62.2$ & $37.7$ & $35.1$ & $81.4$ & $56.9$ & $57.0$ & $81.6$ \\
\hline
SimCSE & $74.0$ & $58.5$ & $75.2$ & $73.6$ & $81.6$ & $55.5$ & $61.1$ & $82.9$ \\
ESimCSE & $73.4$ & $61.2$ & $76.2$ & $74.9$ & $82.9$ & $57.8$ & $57.9$ & $82.1$ \\
SCL & $\bf 78.9$ & $\bf 66.2$ & $\bf 77.8$ & $75.4$ & $\bf84.1$ & $\bf63.3$ & $\bf64.2$ & $\bf84.1$ \\
\hline
& & & & & & & & \\
\hline
& \multicolumn{2}{c}{GoogleNews-T} & \multicolumn{2}{c}{GoogleNews-S} & \multicolumn{2}{c}{GoogleNews-TS} & \multicolumn{2}{c}{Averge} \\
 & ACC & NMI & ACC & NMI & ACC & NMI & ACC & NMI \\
% \cline { 2 - 9} 
\hline
SBERT & $54.9$ & $77.8$ & $58.4$ & $79.9$ & $66.5$ & $86.6$ & $58.8$ & $61.9$ \\
WordNet & $61.2$ & $81.3$ & $67.7$ & $\bf85.5$ & $71.9$ & $90.0$ & $70.0$ & $74.3$\\
Back-Trans & $60.8$ & $80.7$ & $67.1$ & $85.0$ & $72.2$ & $89.8$ & $71.1$ & $76.5$\\
Contextual & $62.1$ & $81.9$ & $66.7$ & $84.6$ & $73.4$ & $\bf90.2$ & $64.7$ & $70.4$\\
\hline
SimCSE & $60.1$ & $80.3$ & $63.8$ & $83.6$ & $71.4$ & $89.1$ & $69.6$ & $74.8$\\
ESimCSE & $ 62.7$ & $82.0$ & $65.2$ & $84.9$ & $72.0$ & $89.8$ & $70.0$ & $76.1$\\
SCL & $\bf 63.2$ & $\bf 82.2$ & $\bf 68.8$ & $85.1$ & $\bf 74.3$ & $89.6$ & $\bf 72.9$ & $\bf 78.0$\\
\hline
\end{tabular}
    \caption{\centering Clustering results on seven datasets with six text augment methods. The first four methods build true positive samples based on extra information and  the last three methods build virtual positive samples without any extra information.}
    \label{table-baseline} \centering
\end{center}
\end{table*}

\textbf{Dataset:} We compare the performance achieved by our SCL with other existing methods on seven benchmark datasets for the short text clustering task. Table \ref{table-dataset} provides an overview of their main statistics. The details of these datasets are provided as follows:
\begin{itemize}

    \item \textbf{SearchSnippets} dataset contains 12,340 snippets extracted from the web search snippets \cite{phan2008learning}, which are categorized into 8 classes.
    
    \item \textbf{StackOverflow} dataset is a subset of the challenge data published by Kaggle, and contains 20,000 question titles selected by \cite{xu2017self} of 20 categories.

    \item \textbf{AgNews} dataset is a subset of news titles \cite{zhang2015text}, which contains 8000 news of 4 categories selected by \cite{rakib2020enhancement}
    
    \item \textbf{Tweet} dataset contains 472 tweets \cite{yin2016model} of 89 categories.
    
    \item \textbf{GoogleNews} contains 152 categories with titles and snippets of 11,109 news articles\cite{yin2016model}. Following \cite{rakib2020enhancement, zhang2021supporting}. Specifically, it contains three sub-set, including GoogleNews-S containing news from snippets, GoogleNews-T containing news from titles and GoogleNews-TS containing news from both snippets and titles.
    
    %we also name the full dataset as GoogleNews-S, and GoogleNews-T and GoogleNews-TS.
    
\end{itemize}

\textbf{Implementation details:} We choose "distilbert-base-nli-stsb-mean-tokens" as the encoder for transformer-based language models, and "glove.6B.300d" as the encoder for word embeddings.  Adam is employed as the optimizer with the learning rate of 5e-5, beta 1 of 1.0 and beta 2 of 1e-4. The dropout rate used in our model is set as 0.3 for all dropout layers. We employ the Accuracy (ACC) and Normalized Mutual Information(NMI) as the evaluation metrics.

\begin{figure*}[t]
	\centering
	\includegraphics[width=400px]{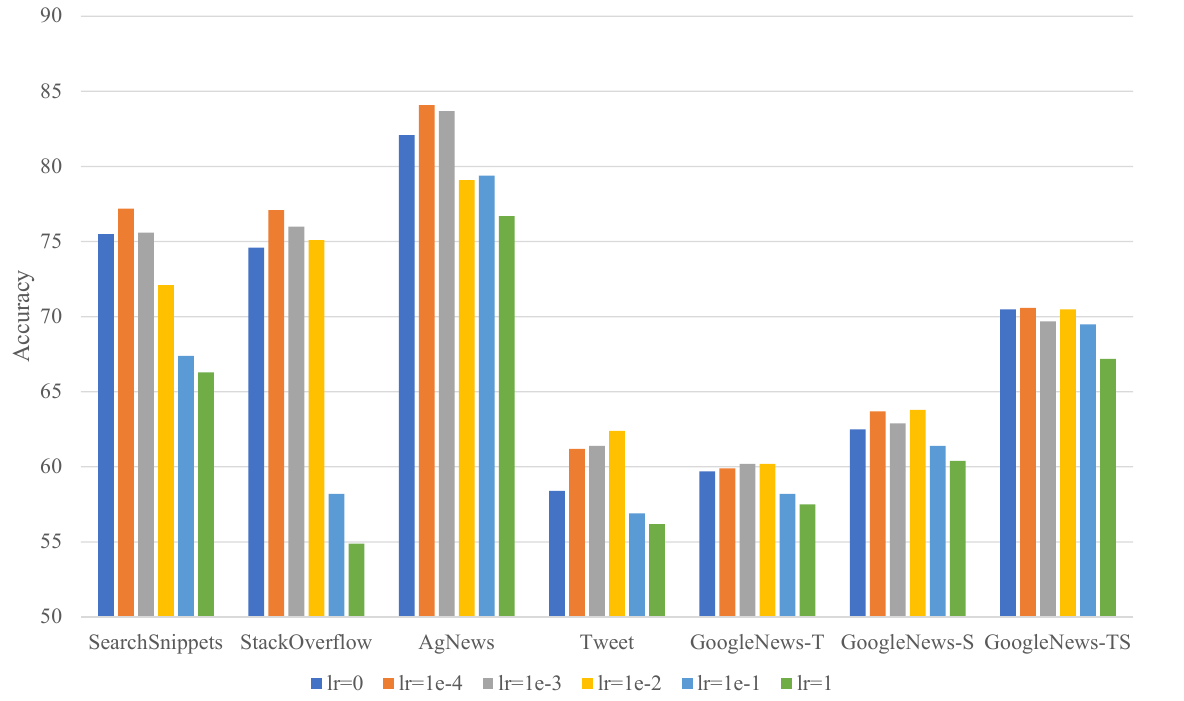}
	\caption{\label{figure-lr} The influence of the parameter $\lambda_{reg}$ range from 1e-4 to 1. A slight $\lambda_{reg}$ can gain better performance compared with no regularization or two much regularization.}
\end{figure*}

\subsection{Results and analysis}
\label{Results and analysis}

\subsubsection{Comparison with existing contrastive learning methods}

To show the superiority of the proposed approach, Table \ref{table-baseline} compares it with multiple existing contrastive learning methods which are listed as follows:
\begin{itemize}
    \item \textbf{SBERT} \cite{DBLP:journals/corr/abs-1908-10084} is the baseline backbone, which is a transformer-based model pre-trained on the combination of the SNLI \cite{BowmanAPM15} and the Multi-Genre NLI \cite{WilliamsNB18} dataset. Then, the model is tuned on three supervised objects. 
    
    \item \textbf{WordNet Augmenter} \cite{MorrisLYGJQ20} substitutes words by the synonyms of WordNet to minimize semantic changes.
    
    \item \textbf{Back translation} \cite{ShenOAR19} generates augmented sentences of the input texts by first translating it to another language (Chinese) and then back to English. 
    
    \item \textbf{Contextual Augmenter} \cite{DBLP:journals/corr/abs-1805-06201} replaces words with other words predicted by pre-training language models according to a context.
    
    \item \textbf{SimCSE} \cite{gao2021simcse} takes dropout as a minimal data augmentation method, and passes the same input sentence to a pre-trained Transformer encoder twice to generate the positive pairs.
    
    \item \textbf{ESimCSE} \cite{wu2021esimcse} is an enhanced version of SimCSE. As the position embedding encoded by transformer relates to the length of input sentences all positive pairs representations in SimCSE incline to be more similar compared with those negative samples with different length. The ESimCSE applies a simple repetition to alleviate the aformentioned issue.

\end{itemize}
As we can see from the table, the proposed SCL achieved the state-of-the-art or comparable performances on all short text clustering datasets in comparison to listed approaches. Specifically, for datasets with fewer categories, SCL can achieve better performances than all other augment methods, i.e., our SCL improved the previous state-of-the-art method NMI from 57.8 to 63.3 on AgNews. Since there are only 4 categories, our approach can sample each category uniformly in every training iteration and generate good virtual positive samples. Meanwhile, the improvements of our approach are not significant on the three GoogleNews datasets, as there are too many categories defined in these dataset, making our model hard to form a discriminative subspace. 

\begin{table*}[t]
\begin{center}
\begin{tabular}{lcccccccc} 
\hline
& \multicolumn{2}{c}{ SearchSnippets } & \multicolumn{2}{c}{ StackOverflow } & \multicolumn{2}{c}{ AgNews } & \multicolumn{2}{c}{ Tweet } \\
& ACC & NMI & ACC & NMI & ACC & NMI & ACC & NMI \\
\cline { 2 - 9 } 
\hline
Glove & $69.8$ & $55.3$ & $23.4$ & $18.5$ & $81.9$ & $58.5$ & $51.9$ & $79.9$ \\
SimCSE & $72.2$ & $58.1$ & $18.6$ & $13.8$ & $82.3$ & $59.4$ & $53.8$ & $80.5$ \\
ESimCSE & $72.6$ & $59.1$ & $17.4$ & $12.2$ & $82.3$ & $59.6$ & $53.9$ & $81.0$ \\
SCL & $\bf80.2$ & $\bf61.0$ & $\bf 27.3$ & $\bf 21.2$ & $\bf 86.3$ & $\bf 62.1$ & $\bf55.1$ & $\bf81.3$ \\
\hline
& & & & & & & & \\
\hline
& \multicolumn{2}{c}{GoogleNews-T} & \multicolumn{2}{c}{GoogleNews-S} & \multicolumn{2}{c}{GoogleNews-TS} & \multicolumn{2}{c}{Averge} \\
 & ACC & NMI & ACC & NMI & ACC & NMI & ACC & NMI \\
\cline { 2 - 9} 
\hline
Glove & $60.6$ & $81.9$ & $57.5$ & $79.9$ & $64.9$ & $86.4$ & $58.6$ & $65.8$ \\
SimCSE & $61.3$ & $82.2$ & $60.6$ & $81.9$ & $67.9$ & $88.3$ & $59.5$ & $66.3$\\
ESimCSE & $63.1$ & $83.5$ & $60.9$ & $81.6$ & $68.3$ & $88.5$ & $59.8$ & $66.5$\\
SCL & $\bf 63.6$ & $\bf 83.6$ & $\bf 61.5$ & $\bf 82.6$ & $\bf 68.7$ & $\bf 89.1$ & $\bf63.2$ & $\bf 68.7$ \\
\hline
\end{tabular}
    \caption{\centering Clustering results based on Glove. }
    \label{table-glove} \centering
\end{center}
\end{table*}

\subsubsection{Subspace affinity}

% clarify if the affinity can represent the cluster-wise relation.
\begin{figure}[h]
	\centering
	\includegraphics[width=250px]{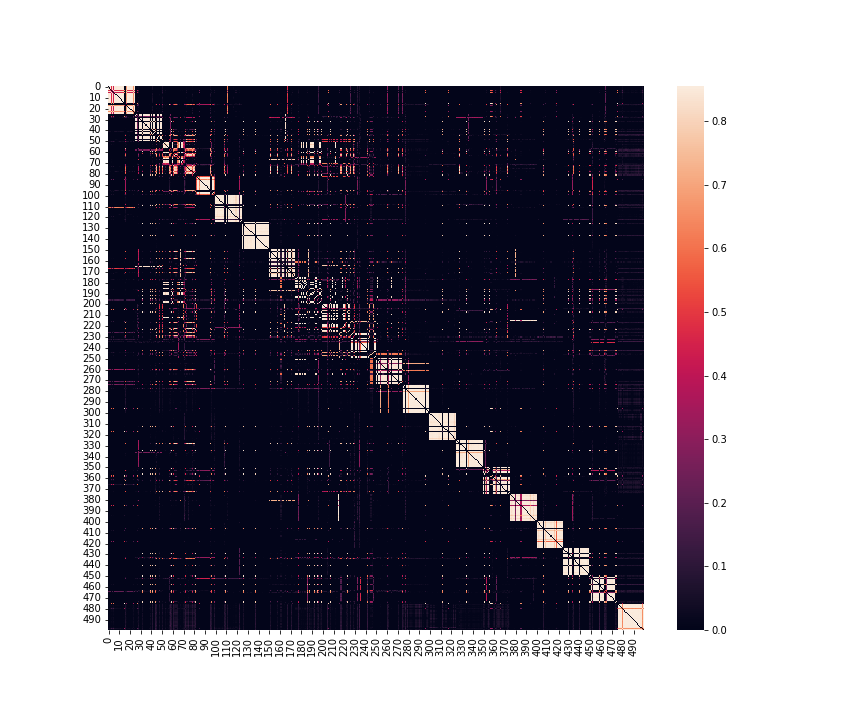}
	\caption{\label{figure-heatmap} The heatmap of subspace affinity matrix on StackOverflows. And brighter points mean higher similarity between samples.}
\end{figure}

The main novelty of our proposed method is the proposed subspace contrastive learning which allows the learned representations to be cluster-aware. We show its advantage in the view of the affinity matrix. In particular, we randomly sample 500 sentences from different categories and feed them into the self-expressive layer to generate the affinity matrix without shuffle. Figure \ref{figure-heatmap} displays the heatmap of an affinity matrix, where the brighter the color displayed, the higher the similarity between the corresponding sentence pair. It can be observed that there are clear white squares appeared in the diagonal of the heatmap, while most of the rest parts are dark. Since pairs represented by the diagonal belongs to the same cluster, the achieved result indicates that the learned subspace can accurately represent the task and cluster-related features from the input texts.

Then, Figure ~\ref{figure-lr} compares the coincidence between those subspaces and clusters, where the regularization parameter $\lambda_{reg}$ is highly correlated with affinity sparsity that directly influences the formation of subspace, and $\lambda_{reg}$ ranges from $1e-4$ to $1$. It can be observed that when $\lambda_{reg}$ is larger than $1e-1$, lower performances were achieved. This is because the larger $\lambda_{reg}$ would lead more punishment on sparsity's loss which leads to the samples only building connections with those most similar samples. Therefore, we need to tune the regularization parameter carefully to obtain a good balance.

\subsubsection{Adapted temperature}
% different temperature and adapted temperature 
% 此处需要画一个折线图，x轴为温度参数，y轴为准确率，并分为有无自适应温度两条线。
% done
\begin{figure}[h]
	\centering
	\includegraphics[width=200px]{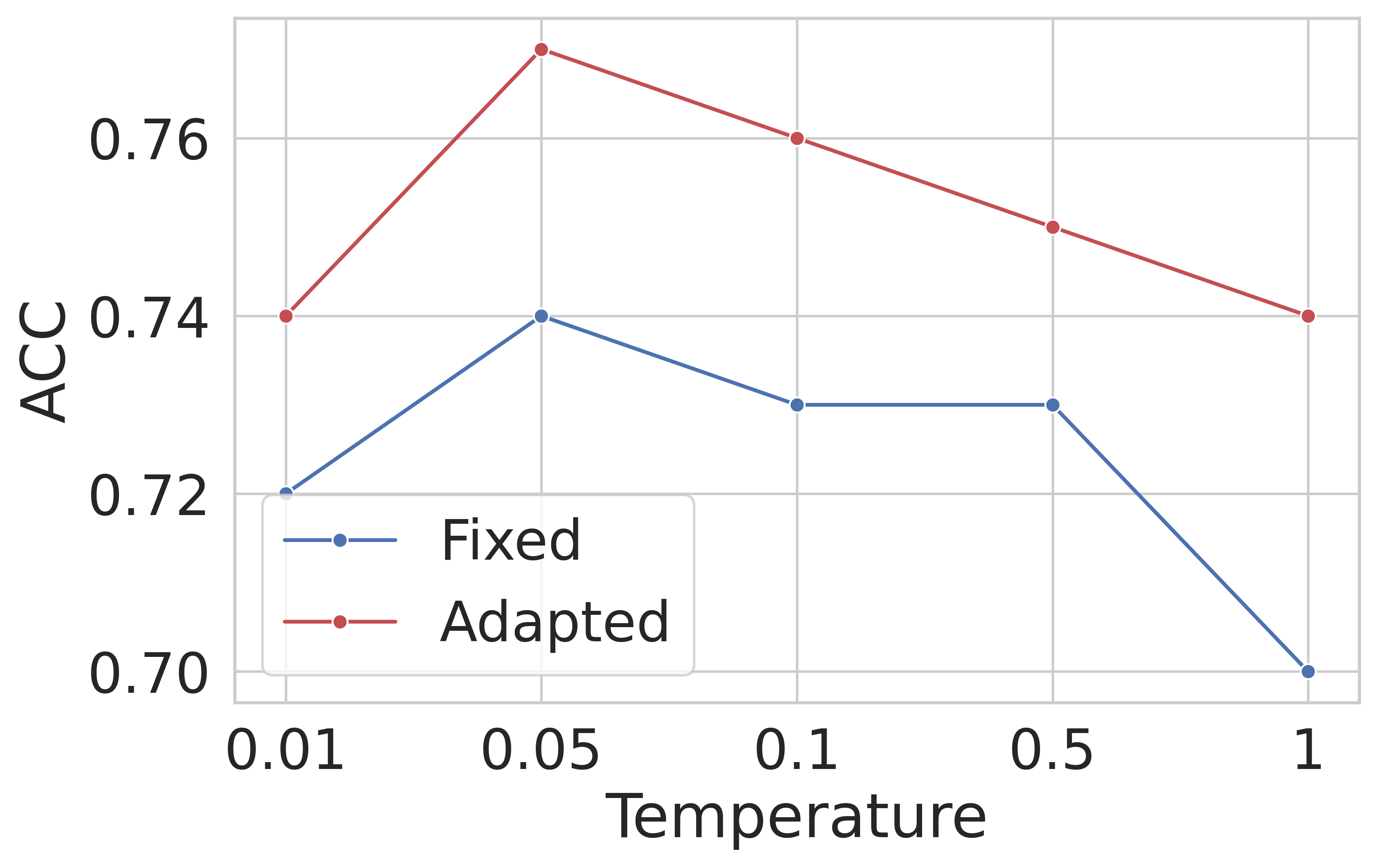}
	\caption{\label{figure-adapted} Grid search of the temperature parameter and the adapted temperature on StackOverflow.}
\end{figure}

Another key parameter in our contrastive learning method is the temperature which plays a role in controlling the strength of penalties on the hard negative samples. Specifically, our contrastive loss with small temperature tends to penalize heavily on the hardest negative samples. As a result, the learned local structure tends to be more discriminative, and the distribution of learned embeddings is likely to be more uniform. Despite that our main optimization loss is still like contrastive loss, its objective functions consist of two aspects. On one hand, the self-expressive layer is trained which simulates the cluster-wise relations and generates virtual positive samples. On the other hand, contrastive learning itself would learn a new sample distribution based on positive samples. In other words, there is a sequential relationship between them, which generates high confidence first and then does contrastive learning on the positive pairs.

We also evaluate the sensitivity of initial temperature setting. As shown in Figure \ref{figure-adapted}, we evaluated initial temperature from 0.01 to 1. It shows that the adapted temperature is stably increased by 0.02 compared with the fixed temperature. We explain this as the adapted temperature can simulate the sequential relationship by raising the temperature in the early stage of training and reducing the temperature as the self-expressive positive samples become more confident.

\subsubsection{Comparison on word embeddings}

% comparison on a dataset with both pre-training glove and bert to demonstrate the meaningful of our positive samples.
% 画一个分组柱状图，x轴为不同的正样本构建方法，y轴为acc。
In comparison with other contrastive learning methods which are mainly based on generating positive samples, our SCL not only has achieved better performance on transformer-based pre-training models, but also generalizes well on shallow networks such as glove. Meanwhile, the traditional text augmentation methods always suffer from a key problem: the sentence semantics is prone to change when replacing and deleting some words in the sentence. Although SimCSE and ESimCSE used dropout as a minimal data augmentation method to address this issue, which is simple and effective,  some information would be lost on shallow networks such as glove. Table \ref{table-glove} compared those methods based on glove. The improvements from SimCSE and ESimCSE decrease because dropout leads to glove having lower similarity compared with transformer-based language models or simple text augments. In contrast, our SCL can consistently provide improvements for various encoder architectures.

% \subsection{Different sample recall methods}
% % 不同类型的召回方法对比
% There is a recondition for subspace clustering methods, every training batch data must have some similar samples so that the subspace can be built. We compare three different data generator methods including: randomly sampling, kmeans recall and locality-sensetive hashing (LSH). 

\section{Conclusion}

In this work, we propose a novel subspace constrastive learning approach for text clustering, which is called SCL. The main advantage of this approach is that it models both instance-wise relationships and cluster-wise relationships among instances, allowing the generated representations to be cluster-aware. More importantly, it can conduct clustering without requiring the number of categories. The experimental results show that the proposed SCL achieved the state-of-the-art or comparable performances on various short text clustering datasets. Moreover, ablation studies show that (1) the proposed SCL can adapt to temperatures, which solves the sequential problem between contrastive objective and self-expressive objective and provide enhanced performance; and (2) the SCL is more suitable for pre-training word embeddings compared with other data augment methods. 

However, there are still two main limitations of our method. Firstly, the performance of our SCL model is sensitive to the training batch size, a small batch size may lead some samples can not find their subspace. Secondly, our method is not suitable to conduct text clustering when the given texts contain too many categories, or some recall work needs to be done before starting the training process. To further address such limitations, our future work will focus on developing better recall methods to help our SCL to meet the precondition and convergence faster.

% \begin{itemize}
%     \item The performances are related to the batch size. Because more samples in one training batch data mean more samples in the same cluster. A sample might not find its subspace if the batch size is too small.
%     \item This method is not really suitable for those datasets with too many categories, or some recall work needs to be done before starting the training process.
% \end{itemize}

% Both the two limitations are based on the recondition for subspace clustering methods, every training batch data must have some similar samples so that the subspace can be built. And we think some good recall methods can help SCL meet the precondition and convergence faster.

\bibliography{anthology,custom}
\bibliographystyle{acl_natbib}

\end{document}